\documentclass[conference]{IEEEtran}
\IEEEoverridecommandlockouts
\usepackage[letterpaper, top=0.75in, bottom=1in, left=0.625in, right=0.625in]{geometry}
\usepackage{booktabs}

\usepackage{cite}
\usepackage{amsmath,amssymb,amsfonts}
\usepackage{algorithmic}
\usepackage{algorithm}
\usepackage{graphicx}
\usepackage{textcomp}
\usepackage{xcolor}
\usepackage{multirow}
\usepackage{comment}
\usepackage{graphicx}
\usepackage{subcaption}

\def\BibTeX{{\rm B\kern-.05em{\sc i\kern-.025em b}\kern-.08em
    T\kern-.1667em\lower.7ex\hbox{E}\kern-.125emX}}

\begin{document}

\title{RadioKMoE: Knowledge-Guided Radiomap Estimation with Kolmogorov–Arnold Networks and Mixture-of-Experts\\
}

\author{\IEEEauthorblockN{Fupei Guo$^*$, Kerry Pan$^\dagger$, Songyang Zhang$^*$, Yue Wang$^\ddagger$, Zhi Ding$^\S$}

\IEEEauthorblockA{$^*$Department of Electrical and Computer Engineering, University of Louisiana at Lafayette, Lafayette, LA 70504, USA\\
$^\dagger$Department of Electrical Engineering and Computer Sciences, University of California, Berkeley, CA 94720, USA\\
$^\ddagger$Department of Computer Science, Georgia State University, Atlanta, GA 30303, USA\\
$^\S$Department of Electrical and Computer Engineering, University of California, Davis, CA 95616, USA
}

\thanks{\textit{Fupei Guo and Kerry Pan contributed equally to this work.}}
}

\maketitle

\begin{abstract}
Radiomap serves as a vital tool for wireless network management and deployment by providing powerful spatial knowledge of signal propagation and coverage. However, increasingly complex radio propagation behavior and surrounding environments pose strong challenges for radiomap estimation (RME).
In this work, we propose a knowledge-guided RME framework that integrates
Kolmogorov–Arnold Networks (KAN) with Mixture-of-Experts (MoE), namely RadioKMoE. 
Specifically, we design a KAN module to predict an initial coarse coverage map, leveraging KAN's strength in approximating physics models and global radio propagation patterns.
The initial coarse map, together with environmental information, drives our MoE network for precise radiomap estimation. Unlike conventional deep learning models, the MoE module comprises expert networks specializing in distinct radiomap patterns to improve local details while preserving global consistency. Experimental results in both multi- and single-band RME demonstrate 
the enhanced accuracy and robustness of the proposed RadioKMoE in radiomap estimation.

\end{abstract}


\section{Introduction}
The rapid deployment of 5G networks and ongoing development of 6G technologies are driving unprecedented demands for high capacity, low latency, and reliable wireless services \cite{1}, which require efficient spectrum optimization and network management. By providing rich information on the spatial distribution of radio spectra across various frequency bands, the radiomap emerges as an effective tool to enable spectrum-efficient resource allocation and wireless network management. Usually, a dense radiomap is estimated from sparse observations collected by deployed sensors and mobile devices \cite{10634040}. However, the expansion of spectrum bands and complex environments render high-fidelity radiomap reconstruction increasingly challenging in practice, motivating the need for more robust and efficient RME approaches.

Early RME approaches focused on interpolation methods such as Kriging \cite{kriging} and model-based interpolation \cite{6872807}, which infer unmeasured locations from spatial correlations. However, 
their dependence on stationarity and isotropy assumptions of radio propagation behavior, together with ignorance of surrounding objects, restricts their generalizability in complex urban environments. 
To overcome these limitations, deep learning (DL) techniques are applied to RME, using urban environments and sparse observations as input. Typical examples include autoencoder-based methods \cite{AE} and RadioUNet \cite{levie2021radiounet}, which employ convolutional architectures to capture complex spatial dependencies and improve predictive accuracy. Despite some successes, these approaches tend to overfit specific layouts, and require sufficient training data, which can be inaccessible in practical applications. 

With the development of machine learning, recent interests focus on RME based on generative learning paradigms, including generative adversarial network (GAN) based models \cite{RME-GAN} and diffusion-driven frameworks \cite{qiu2023irdm,DDPM-RM}, which produce visually realistic reconstructions while capturing radio propagation uncertainty. Several of these methods further warm-start the reconstruction with a coarse prior to guide global structure learning. Typical approaches include using interpolation-based pre-estimates in conditional GAN pipelines \cite{RME-GAN}, deriving priors from geographic or semantic layers, and leveraging physics-driven ray-tracing toolchains \cite{li2024geo2sigmap}. Although these priors can stabilize training and improve large-scale consistency, they often rely on computationally intensive simulations or strong domain-specific assumptions, limiting scalability and portability across different locations and frequency bands. 

\begin{figure}[t]
    \centering
    \includegraphics[width=\linewidth]{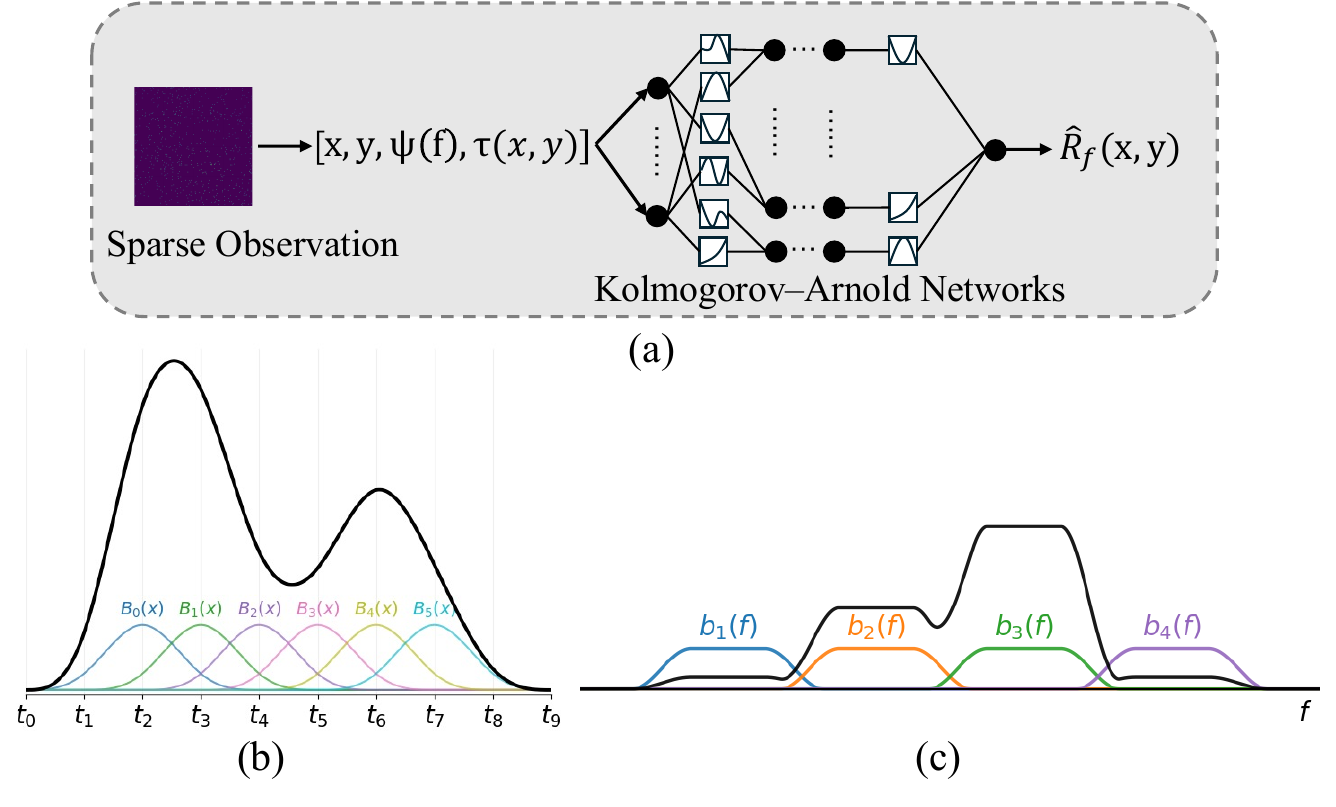}
    \caption{Structure of the Kolmogorov–Arnold Network (KAN): (a) KAN frameworks; (b) Shapes of spline kernels in KAN; (c) BEM representation of multiband spectrum strengths.}
    \label{fig:kan_structure}
    \vspace{-3mm}
\end{figure}

\begin{figure}[t]
    \centering
    \setlength{\tabcolsep}{2pt} 
    \begin{minipage}[b]{0.32\columnwidth}
        \centering
        \includegraphics[width=\linewidth]{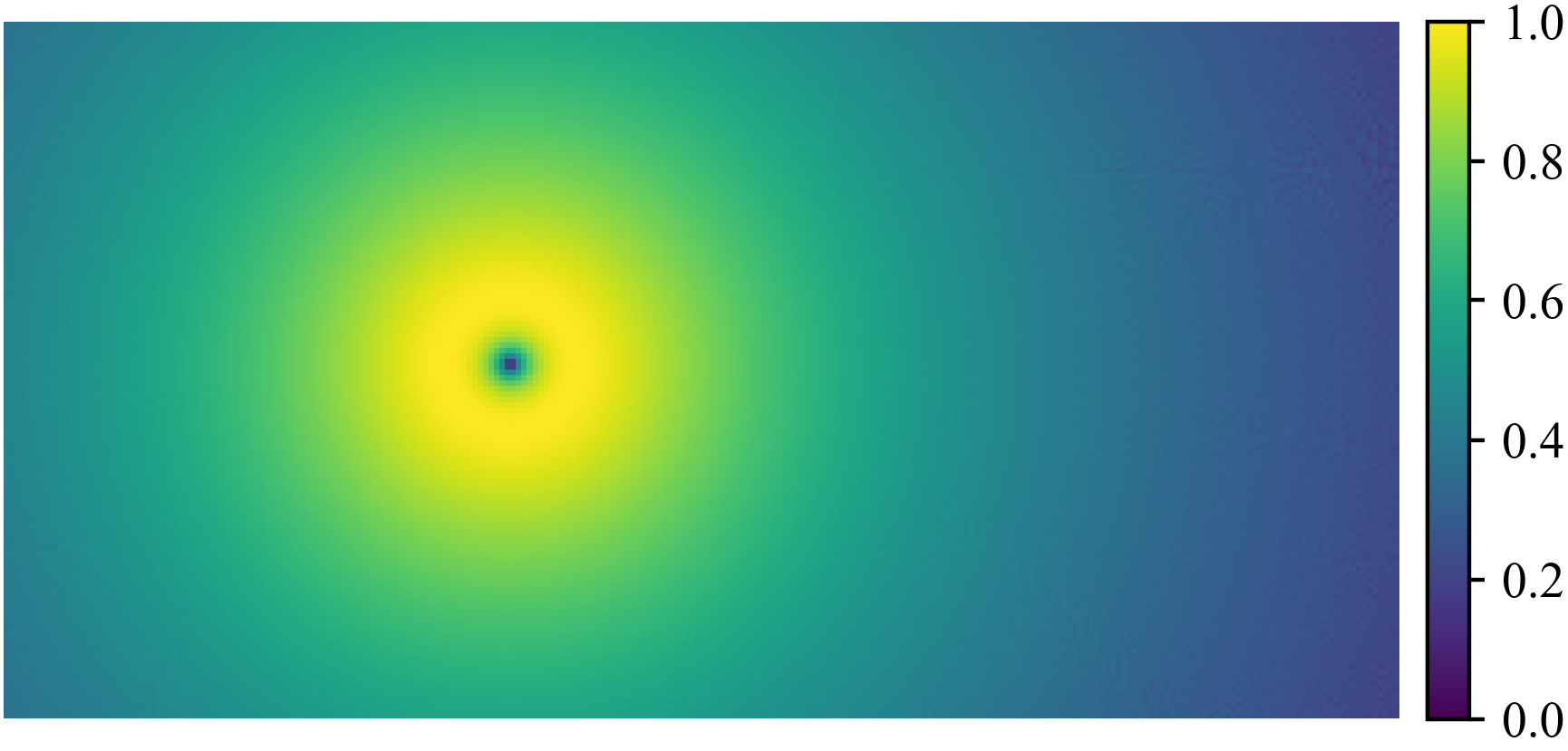}

    \end{minipage}
    \hfill
    \begin{minipage}[b]{0.32\columnwidth}
        \centering
        \includegraphics[width=\linewidth]{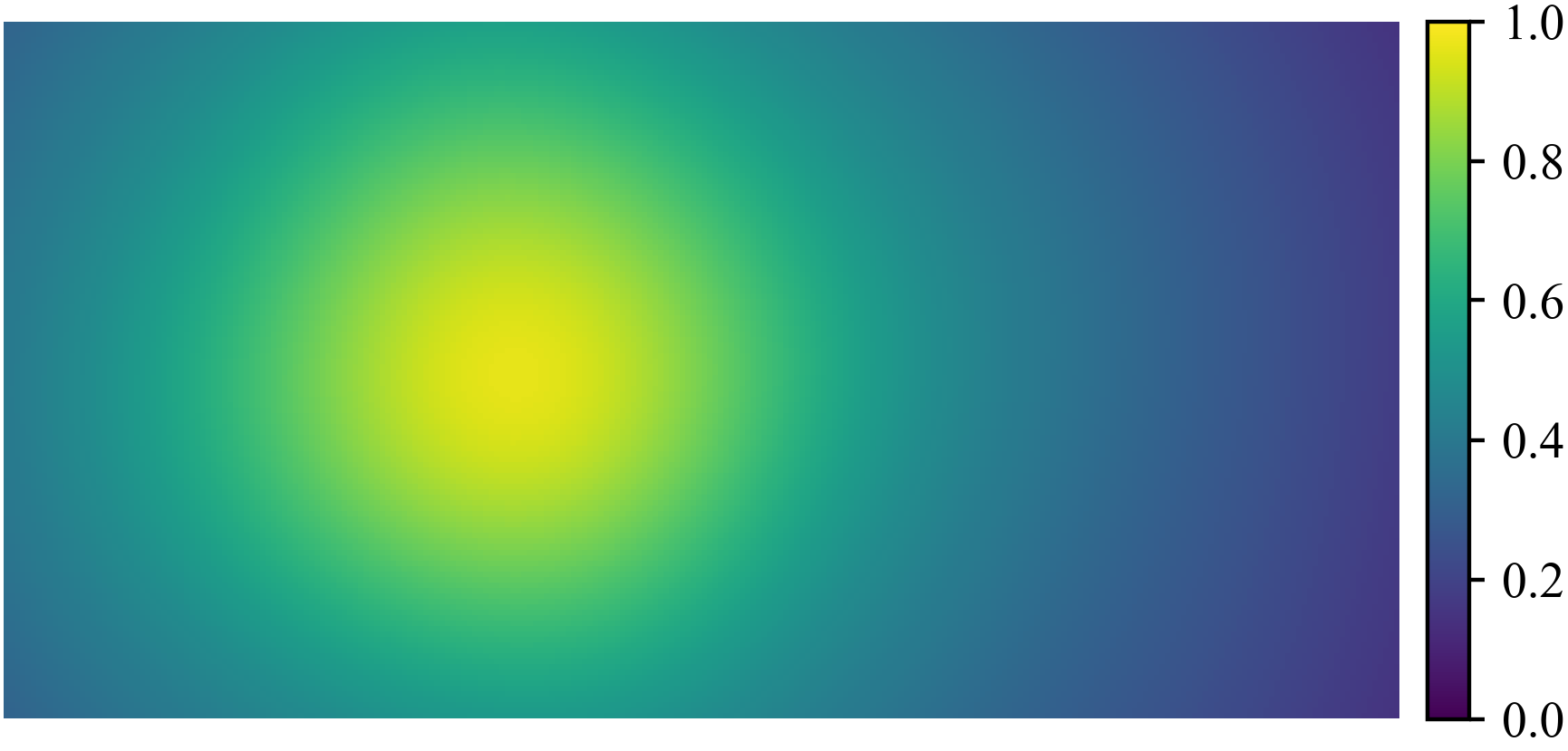}

    \end{minipage}
    \hfill
    \begin{minipage}[b]{0.335\columnwidth}
        \centering
        \includegraphics[width=\linewidth]{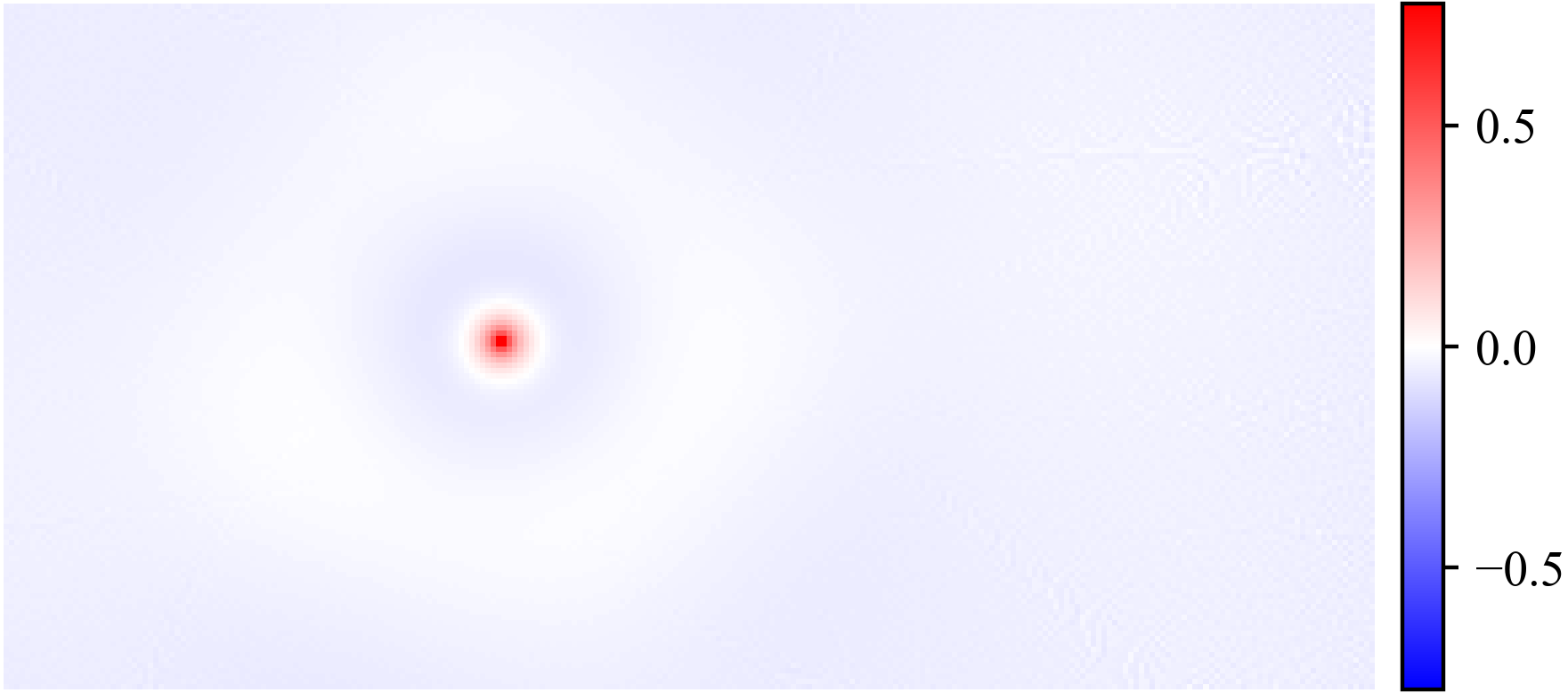}

    \end{minipage}
    \vspace{2mm} 
    \begin{minipage}[b]{0.32\columnwidth}
        \centering
        \includegraphics[width=\linewidth]{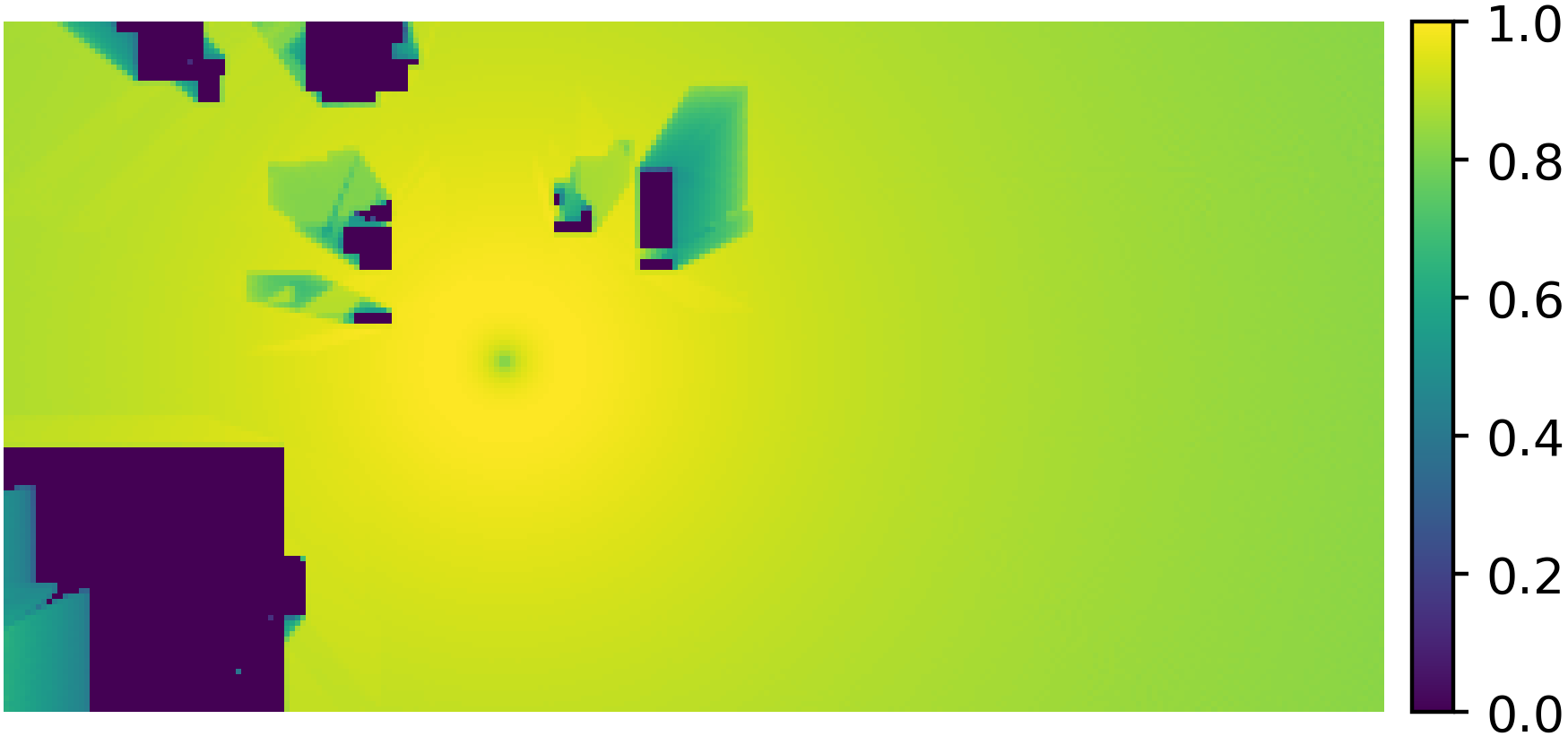}
        \small (a) Ground Truth
    \end{minipage}
    \hfill
    \begin{minipage}[b]{0.32\columnwidth}
        \centering
        \includegraphics[width=\linewidth]{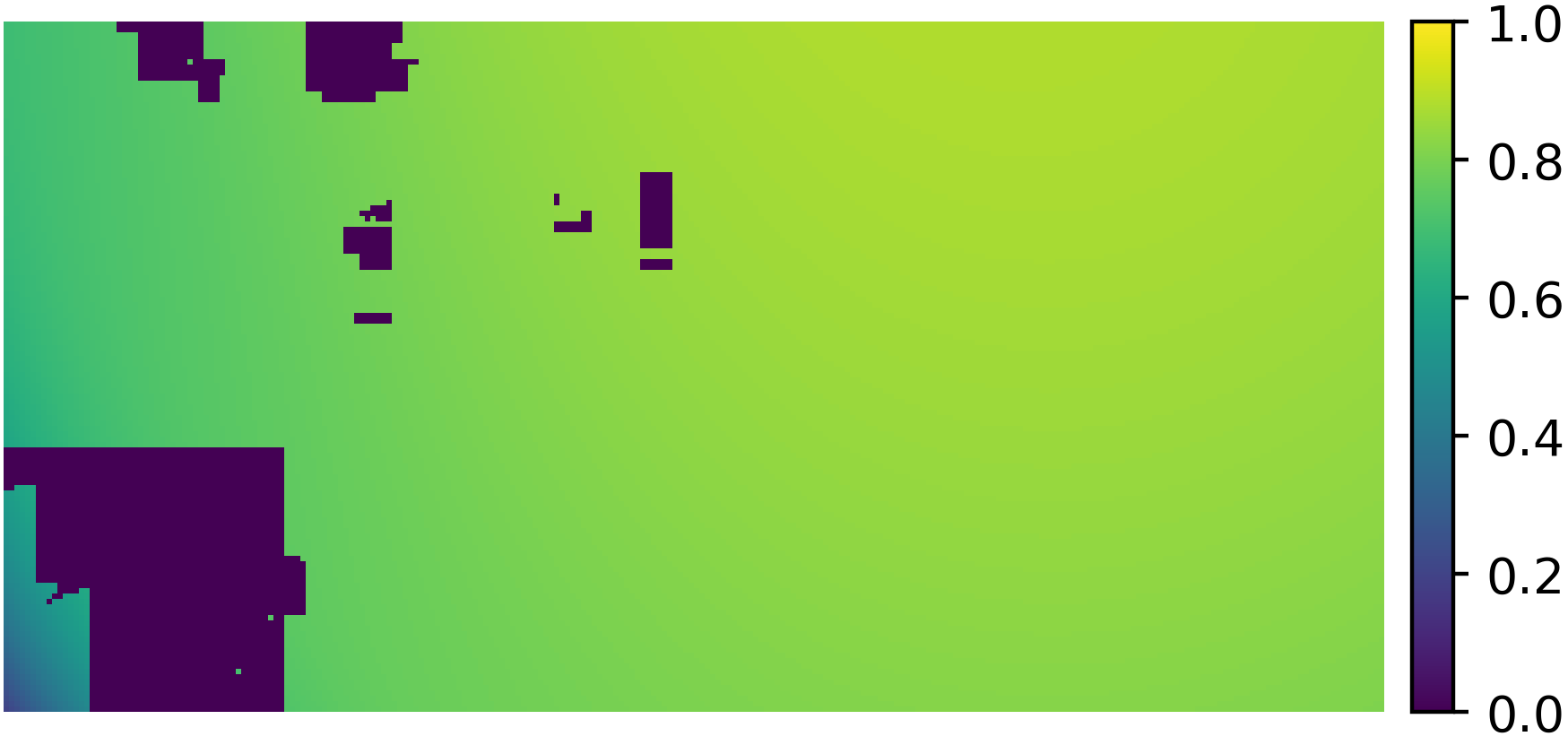}
        \small (b) Prediction
    \end{minipage}
    \hfill
    \begin{minipage}[b]{0.335\columnwidth}
        \centering
        \includegraphics[width=\linewidth]{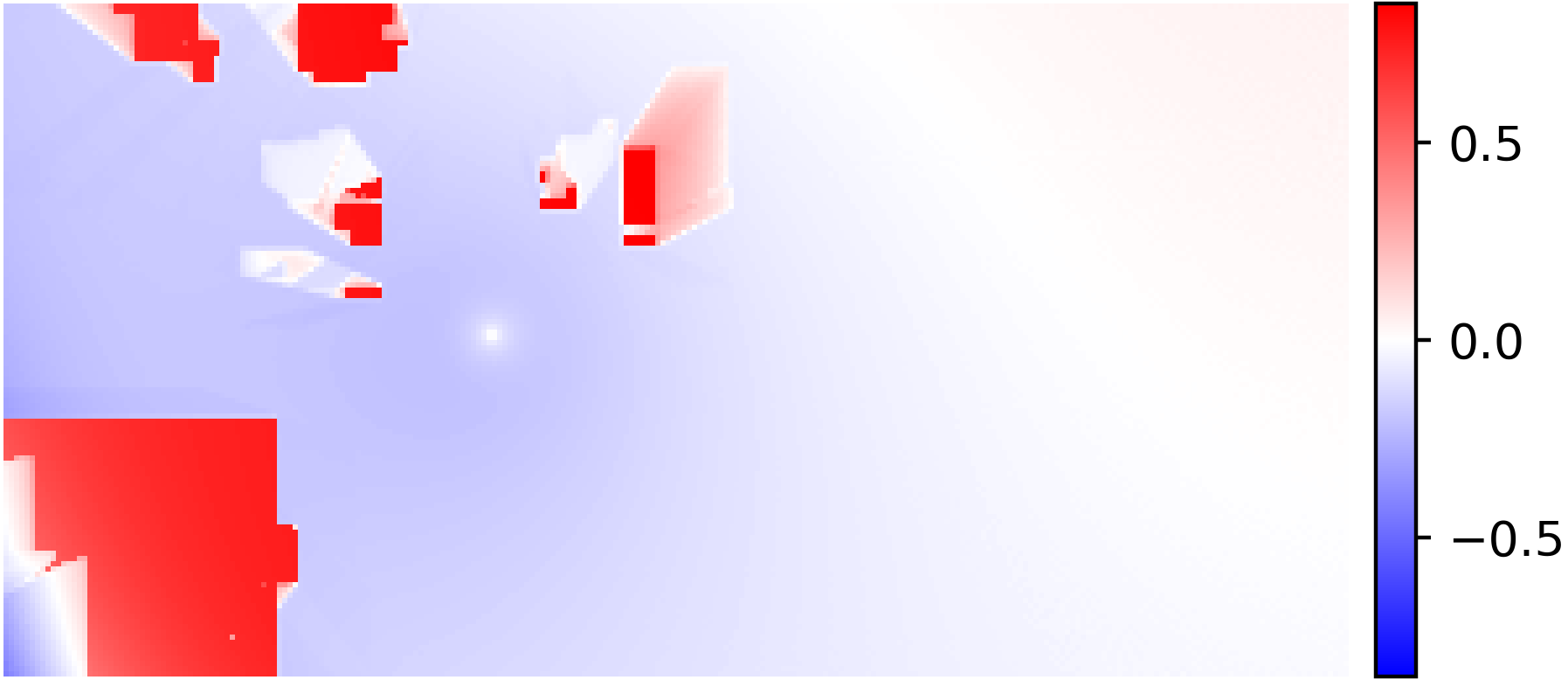}
        \small (c) Error Map
    \end{minipage}

    \caption{KAN prediction results in free-space (top row) and in an urban scene with buildings (bottom row), where complex environments degrade the performance of KAN prediction. }
     \label{fig:kan} 
     \vspace{-5mm}
\end{figure}

To this end, we propose to adopt Kolmogorov–Arnold network (KAN) \cite{liu2025kan} to generate a knowledge-guided prior for RME. Unlike simulators that depend on extensive prior knowledge (material properties, antenna patterns, and detailed geometry) and hand-crafted kernels, KAN learns global propagation regularities directly from data. By replacing fixed activations with edge-wise learnable kernel expansions, it produces smooth, interpretable function approximations with high parameter efficiency. Compared with conventional networks, the KAN architecture provides clearer functional interpretability while maintaining competitive accuracy and efficiency. For example, as shown in Fig. \ref{fig:kan_structure}, the spline kernels naturally fit the basis expansion model (BEM) for multiband spectrum strengths \cite{5934611}. As an emerging technology, KAN-based radiomap estimation remains underexplored. The only existing work is RadioKAN \cite{11143592}, which focuses on single-band RME and ray-tracing. Although KAN is effective in learning smooth, large-scale radio propagation patterns, it struggles to represent sharp spatial variations introduced by obstacles. As shown in Fig.~\ref{fig:kan}, KAN performs well in open areas but oversmooths transitions at building edges, yielding boundary-localized errors and failing to reproduce shadowing patterns. To capture environmental information, one may incorporate KAN with building information. However, its smooth inductive bias still limits its ability to capture abrupt attenuation and multipath effects. This motivates a complementary refinement stage in the KAN prediction to recover localized variations.

Thus, in this work, we introduce a novel knowledge-guided RME framework, namely \textbf{RadioKMoE}, which integrates KAN networks and MoE \cite{shazeer2017outrageously}. Specifically, the KAN is applied to produce a fast coarse coverage estimate that preserves global trends, which is then concatenated with environmental cues and a physics-inspired radio depth map \cite{brat1} as input for an MoE module. The MoE module tokenizes the priors and routes each token to specialized expert networks (e.g., characterizing various open areas and building-dense blocks), enabling adaptive, fine-grained modeling \cite{Riquelme2021VisionMoE,Dryden2022SpatialMoE}. 
Leveraging KAN's ability to capture global radio propagation behavior, together with the power of MoE in adaptively refining local regions through expert specialization,
the proposed RadioKMoE enables accurate and scalable RME under diverse urban conditions with multiple frequency bands. 

Our contribution can be summarized as follows:

\begin{itemize}
    \item We propose a novel knowledge-guided framework (RadioKMoE) that unifies global propagation modeling with KAN networks and localized refinement with MoE for effective radio map estimation.
    \item We incorporate a physics-inspired radio depth map with MoE refinement, enabling adaptive modeling of heterogeneous urban environments and enhancing fine-grained reconstruction quality guided by physical radio propagation behavior.
    \item Our framework is capable of addressing both single-band and multiband RME, which enhances the generalizability of KAN-based RME.
\end{itemize}


\begin{figure*}[t]
    \centering
    \includegraphics[width=0.85\textwidth]{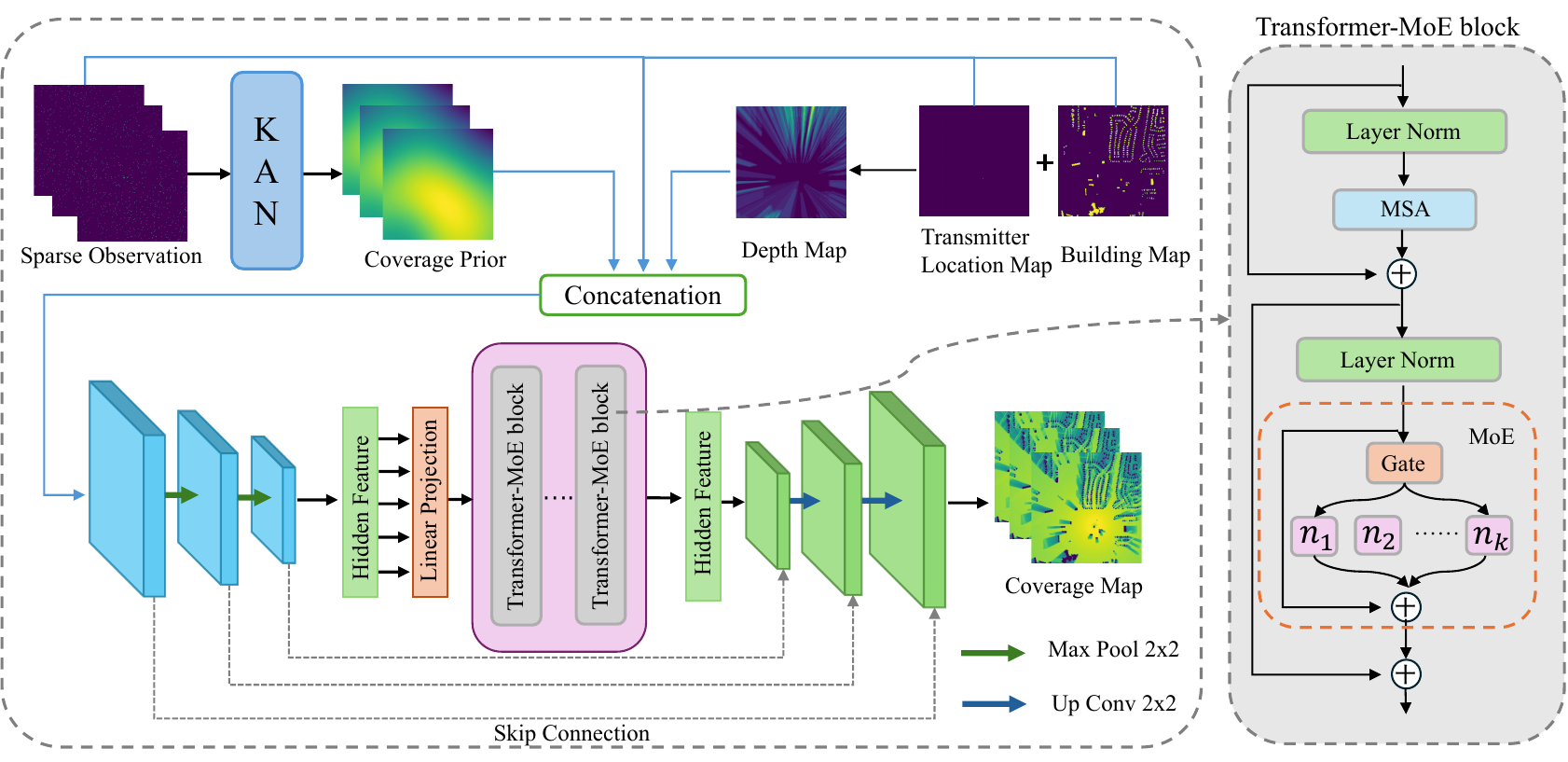}
    \caption{Overview of the proposed RadioKMoE architecture. Stage 1: Coarse resolution radiomap is generated by KAN as priors. Stage 2: The knowledge-guided priors (KAN-based coverage priors, radio depth map) and environment priors (building map, transmitter locations) are input into a MoE-based transformer for refinement and final radiomap estimation.}
    \label{fig:model}
\end{figure*}
\section{Problem Formulation}
In this work, we focus on multiband radiomap estimation in complex urban environments from sparse observations.


Let $\mathcal{A} \subset \mathbb{R}^2$ denote the target geographical area, discretized into an $H \times W$ grid (e.g., $256 \times 256$). For each transmitter location $t \in \mathcal{T}$ and frequency $f \in \mathcal{F}$, the radio map is represented as
\(
R_{f}(x,y) = P_{t,f}(x,y), \quad (x,y) \in \mathcal{A}
\),
where $P_{t,f}(x,y)$ denotes the received power (or path-loss) at position $(x,y)$. The set of frequencies is denoted by $\mathcal{F} = \{ f_1, f_2, \dots, f_M \}$, where $M$ is the number of bands.

In practice, the radiomap $R_f(x,y)$ is only partially observed due to limited measurements. We denote the observation set as
\begin{equation}
\mathcal{O}_f = \{ \big((x_i,y_i), R_f(x_i,y_i)\big) \mid i = 1, \dots, N \},
\end{equation}
where $N \ll H \times W$ is the number of sparse measurements. 

The objective is to reconstruct the full coverage map $\hat{R}_f(x,y)$ for all $(x,y) \in \mathcal{A}$ and $f \in \mathcal{F}$, given the sparse observations $\mathcal{O}_f$ and the environmental information.

To model the effect of the environment, we define an auxiliary map $E(x,y)$ that encodes static obstacles such as buildings and open spaces. $E(x,y) = 1$ if $(x,y)$ is occupied by a building and $E(x,y) = 0$ otherwise. In addition, we incorporate a depth map $D(x,y)$ that encodes the relative distance to transmitters, providing global context for signal attenuation. The problem can be formulated as
\begin{equation}
\{\hat{R}_f|f\in \mathcal{F}\} = \mathcal{P}_{\theta} \big( \{\mathcal{O}_f|f\in \mathcal{F}\}, E, D\big),
\end{equation}
where $\mathcal{P}_{\theta}$ denotes the mapping function parameterized by $\theta$. The goal is to minimize the reconstruction error across all frequencies and spatial positions, denoted by
\begin{equation}
\min_{\theta} \sum_{f \in \mathcal{F}} \sum_{(x,y) \in \mathcal{A}} 
\ell\big( \hat{R}_f(x,y), R_f(x,y) \big),
\end{equation}
where $\ell(\cdot,\cdot)$ characterizes reconstruction error (e.g., mean squared error or normalized mean square error).  


\section{Method}
In this section, we present a knowledge-guided framework that couples a KAN for fast global prediction with a MoE for fine-grained refinement. The pipeline operates in two stages as shown in Fig. \ref{fig:model}. 
Firstly, a KAN module ingests multi-frequency sparse observations together with the operating frequency to produce an initial coverage prior. 
Secondly, the KAN-based coverage is fused with environmental cues (e.g., building/layout maps) and physics-inspired radio depth as input to an MoE network that adaptively assigns region-wise experts to refine local details. This design leverages KAN's efficiency for global propagation modeling and MoE's specialization for heterogeneous urban environments.

\subsection{Knowledge-guided Coverage Prior}
\label{subsec:kan}
In Stage 1, we employ KAN to construct the knowledge-guided prior, leveraging its ability to provide an efficient and interpretable function approximation. 
Unlike traditional MLPs that rely on fixed nonlinear activations, KAN replaces them with learnable univariate functions on edges.

Suppose that $x^{(\ell)}\!\in\!\mathbb{R}^{n_\ell}$ denotes the activations at layer $\ell$, a KAN layer computes
\begin{equation}
x^{(\ell+1)}_j \;=\; \sum_{i=1}^{n_\ell} \varphi^{(\ell)}_{j,i}\!\big(x^{(\ell)}_i\big), 
\qquad j=1,\dots,n_{\ell+1}.
\label{eq:kan-layer}
\end{equation}
where each edge function $\varphi^{(\ell)}_{j,i}:\mathbb{R}\!\to\!\mathbb{R}$ is learnable. 
A practical parameterization expands every $\varphi^{(\ell)}_{j,i}$ on a compact basis (e.g., splines) with coefficients trained end-to-end, which yields smooth, data-adaptive nonlinearities while keeping the number of parameters modest. 
By composing Eq. \eqref{eq:kan-layer} across layers, KAN efficiently models smooth, globally coherent fields—an inductive bias that aligns with overall radio propagation.

\subsubsection{KAN for coarse interpolation}
We model the coverage estimation as a pointwise regression function learned by KAN. 
For each frequency $f\in\mathcal{F}$ and location $(x,y)\in\mathcal{A}$, we form a feature vector
\(
\boldsymbol{\phi}(x,y)
=\big[x,\;y,\;\{\Psi(f)|f\in\mathcal{F}\},\;\tau(x,y)\big]
\),
where $\Psi(f)$ is a scalar or low-dimensional embedding of the carrier, and $\tau(x,y)$ stacks optional factors (e.g., transmitter-location indicators, distance-to-Tx, or other side cues available at $(x,y)$). 
KAN implements a learnable mapping $G_{\phi}:\mathbb{R}^{d}\!\to\!\mathbb{R}$ that predicts the received power (or path loss) at a single coordinate denoted by
\begin{equation}
\widehat{R}(x,y) \;=\; G_{\phi}\!\big(\boldsymbol{\phi}(x,y)\big).
\label{eq:kan-pointwise}
\end{equation}
Given the learned $G_{\phi}$, the coarse coverage prior over the entire region is obtained by evaluating Eq. \eqref{eq:kan-pointwise} at every location for each frequency, i.e.,
\begin{equation}
\widehat{R} \;=\; 
\big\{\widehat{R}_f(x,y)\,|\,(x,y)\in\mathcal{A},f\in\mathcal{F}\big\}\in\mathbb{R}^{H\times W\times F}.
\end{equation}

\subsubsection{Training objective}
Let $\mathcal{O}_f=\{((x_i,y_i),R_f(x_i,y_i))\}_{i=1}^{N_f}$ be a set of $N_f$ observed samples for band $f$. 
We fit $G_{\phi}$ by minimizing the prediction error, optionally complemented with a spatial regularizer that encourages smooth, physically plausible fields, calculated by
\begin{align}
\mathcal{L}_{\mathrm{KAN}}(\phi)
=\sum_{f\in\mathcal{F}}\sum_{i=1}^{N_f}
\Big(\widehat{R}_f(x_i,y_i)-R_f(x_i,y_i)\Big)^{2}.
\end{align}
This pointwise formulation lets KAN learn a coordinate-conditioned function that generalizes across the domain, where evaluating it densely yields the coarse map used as input to the MoE refinement in Sec.~\ref{subsec:moe}.

\subsection{Extraction of Knowledge-Guided Priors}
\label{subsec:cond}
We construct a multi-channel prior tensor by aggregating four knowledge-guided and environment-specified complementary cues as follows:

\textbf{Building map.}
Let $E \in \{0,1\}^{H\times W}$ denote the static layout; $E(x,y)=1$ indicates the presence of a building at grid cell $(x,y)$ and $E(x,y)=0$ denotes open space.

\textbf{Transmitter-location map.}
Let $T \in \{0,1\}^{H\times W}$ encode transmitter positions. For a single transmitter at location $t=(x_t,y_t)$, we set $T(x_t,y_t)=1$ and $T(x,y)=0$ elsewhere. For multiple transmitters $\{t_m\}_{m=1}^{N_t}$, we define $T(x,y)=\min\!\big(1, \sum_{m=1}^{N_t}\mathbf{1}\{(x,y)=t_m\}\big)$.

\textbf{Depth map.}
For each receiver cell $(x,y)$ and transmitter location $t=(x_t,y_t)$, we define a discrete line segment $\mathcal{L}(x,y;t)$ connecting $(x,y)$ and $t$, rasterized using the Bresenham algorithm.
We then count the number of building pixels intersected by this path:
\begin{equation}
D(x,y;t)=\sum_{(u,v)\in\mathcal{L}(x,y;t)} E(u,v).
\end{equation}
To obtain a normalized depth map, we clip the maximum intersection count by a constant $\tau_{\max}$ (e.g., $150$ in our implementation) and scale the result to $[0,1]$:
\begin{equation}
D_\text{n}(x,y;t)=\min\left(1,\frac{D(x,y;t)}{\tau_{\max}}\right).
\end{equation}
The radio depth map generally captures the shadowing patterns and distance impacts.

\textbf{Coarse coverage prior.}
The KAN module produces a coarse full-field estimate $\widehat{R} \in \mathbb{R}^{H\times W\times F}$ (Sec.~\ref{subsec:kan}). We optionally normalize $\widehat{R}_f$ per frequency (e.g., min--max to $[0,1]$) to stabilize subsequent fusion.

\textbf{Prior tensor.}
All feature maps are concatenated along the channel dimension to form the conditioning input for the MoE refinement network. 
Specifically, we combine the coarse coverage prior multiband $\{\widehat{R}_f|f\in\mathcal{F}\}$, the building map $B$, the transmitter-location map $T$, the normalized depth map $N$, and the sparse observation map $S_f$ as
\begin{equation}
\widetilde{\mathbf{C}} = \mathrm{concat}\!\big(\{\widehat{R}_f,S_f|f\in\mathcal{F}\},\, E,\, T,\, D_\text{n} \big) \in \mathbb{R}^{H\times W\times d_{\mathrm{in}}}.\nonumber
\end{equation}

\subsection{MoE-based Refine Model}
\label{subsec:moe}
In Stage 2, we design the refinement network following a TransUNet-style encoder–decoder architecture, augmented with Transformer blocks whose feed-forward layers are replaced by a sparse MoE module. 
The overall process consists of three main components: a UNet encoder for feature extraction, a Transformer–MoE block for adaptive refinement, and a UNet decoder for RME.

\textbf{UNet encoder.}
Given the prior tensor $\widetilde{\mathbf{C}} \in \mathbb{R}^{H\times W\times d_{\mathrm{in}}}$, the encoder extracts multi-scale contextual features through a sequence of convolutional layers with downsampling. 
Each stage captures spatial patterns such as building structures and transmitter layouts while progressively increasing the receptive field. 
The resulting feature maps are flattened and embedded into $K$-dimensional tokens for the Transformer–MoE stage:
\begin{equation}
\mathbf{Z}^{(0)}=\mathrm{PatchEmbed}\!\big(\widetilde{\mathbf{C}}\big)\in\mathbb{R}^{N\times K},\quad 
N=\frac{HW}{P^2},
\end{equation}
where $P$ is the patch size and $N$ the total number of tokens.

\textbf{Transformer–MoE block.}
In this stage, spatial dependencies and semantic relationships are refined through self-attention and expert specialization. 
A standard Transformer layer applies multi-head self-attention (MSA) followed by a feed-forward network (FFN). 
We replace the FFN with a sparse MoE to enable region-wise adaptation. 
For the $\ell$-th block, the computations are
\begin{align}
\mathbf{H}^{(\ell)} &= \mathrm{MSA}\!\big(\mathrm{LN}(\mathbf{Z}^{(\ell)})\big) + \mathbf{Z}^{(\ell)}, \nonumber\\
\mathbf{Z}^{(\ell+1)} &= \mathrm{MoE}\!\big(\mathrm{LN}(\mathbf{H}^{(\ell)})\big) + \mathbf{H}^{(\ell)}.
\end{align}

Each token $\mathbf{h}\in\mathbb{R}^D$ is routed to a small subset of experts selected by a learned router. 
The router computes
\begin{equation}
\boldsymbol{\alpha}=\mathbf{W}_r\,\mathbf{h}+\mathbf{b}_r,\qquad
\mathbf{p}=\mathrm{softmax}(\boldsymbol{\alpha}),
\end{equation}
and activates the top-$k$ experts $\mathcal{S}_k(\mathbf{h})$.
Each expert $e$ is a lightweight position-wise network:
\begin{equation}
\mathrm{Expert}_e(\mathbf{h}) = \mathbf{W}^{(2)}_{e}\,\sigma\!\big(\mathbf{W}^{(1)}_{e}\mathbf{h}+\mathbf{b}^{(1)}_{e}\big)+\mathbf{b}^{(2)}_{e},
\end{equation}
and the final MoE output is the weighted sum of outputs from selected experts, i.e.,
\begin{equation}
\mathrm{MoE}(\mathbf{h}) = \sum_{e\in \mathcal{S}_k(\mathbf{h})} \tilde{p}_e\,\mathrm{Expert}_e(\mathbf{h}),
\qquad 
\tilde{p}_e = \frac{p_e}{\sum_{j\in \mathcal{S}_k(\mathbf{h})}p_j}.
\end{equation}
This formulation allows each expert to specialize in distinct regional patterns in the hidden space (e.g., building-dense, open, or transition areas), improving adaptivity while maintaining computational efficiency.

\textbf{UNet decoder.}
The decoder mirrors the encoder structure and employs skip connections to fuse multi-scale features for reconstruction.
It progressively upsamples the refined tokens from the Transformer–MoE block back to the original spatial resolution, generating a residual coverage map through deconvolutional layers and a final projection.
This design preserves the structural information captured by the encoder while enabling localized refinement near building edges, occlusions, and high-gradient regions.

\begin{figure*}[t]
    \centering
    \setlength{\tabcolsep}{1pt}  
    \renewcommand{\arraystretch}{0.85} 

    \begin{tabular}{ccccccccc}
        \includegraphics[width=0.10\textwidth]{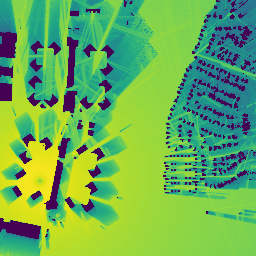} &
        \includegraphics[width=0.10\textwidth]{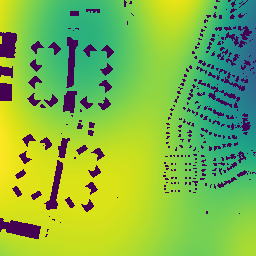} &
        \includegraphics[width=0.10\textwidth]{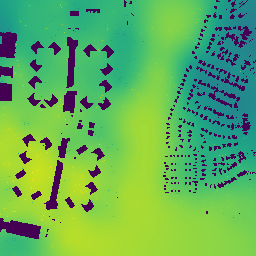} &
        \includegraphics[width=0.10\textwidth]{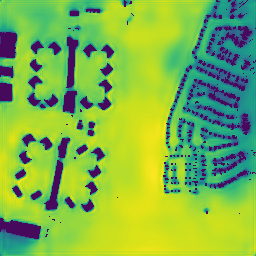} &
        \includegraphics[width=0.10\textwidth]{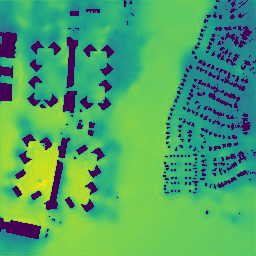} &
        \includegraphics[width=0.10\textwidth]{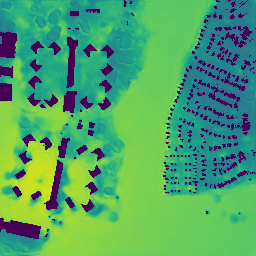} &
        \includegraphics[width=0.10\textwidth]{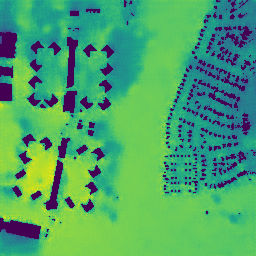} &
        \includegraphics[width=0.10\textwidth]{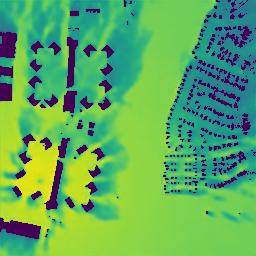} &
        \includegraphics[width=0.10\textwidth]{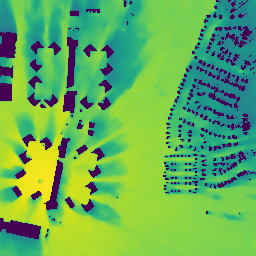} \\[-2pt]
    \end{tabular}

    \vspace{1mm}

    \begin{tabular}{ccccccccc}
        \includegraphics[width=0.10\textwidth]{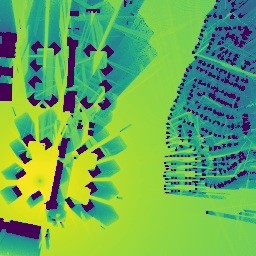} &
        \includegraphics[width=0.10\textwidth]{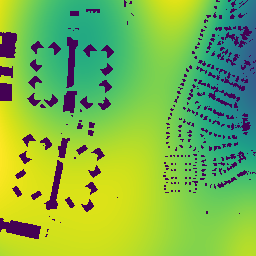} &
        \includegraphics[width=0.10\textwidth]{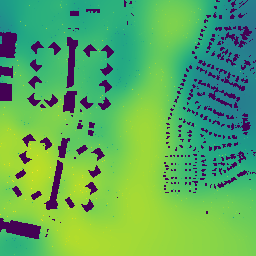} &
        \includegraphics[width=0.10\textwidth]{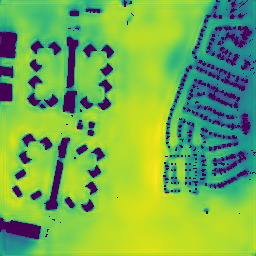} &
        \includegraphics[width=0.10\textwidth]{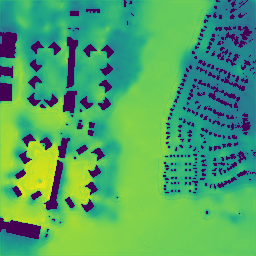} &
        \includegraphics[width=0.10\textwidth]{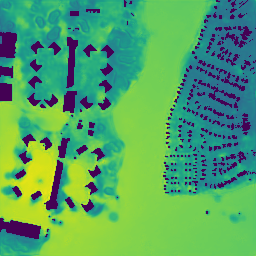} &
        \includegraphics[width=0.10\textwidth]{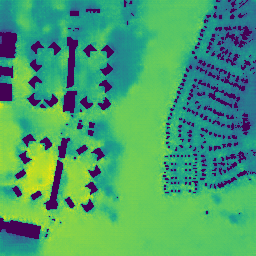} &
        \includegraphics[width=0.10\textwidth]{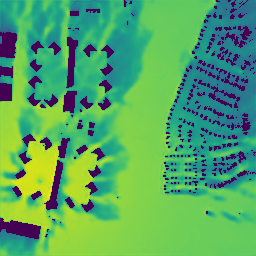} &
        \includegraphics[width=0.10\textwidth]{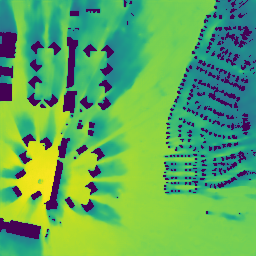} \\[-2pt]
    \end{tabular}

    \vspace{1mm}

    \begin{tabular}{ccccccccc}
        \includegraphics[width=0.10\textwidth]{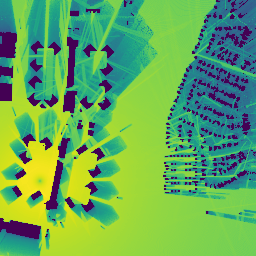} &
        \includegraphics[width=0.10\textwidth]{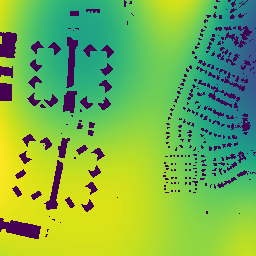} &
        \includegraphics[width=0.10\textwidth]{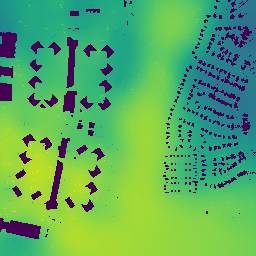} &
        \includegraphics[width=0.10\textwidth]{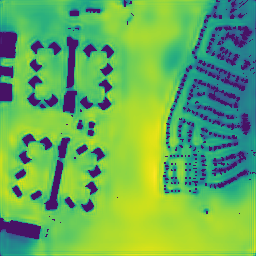} &
        \includegraphics[width=0.10\textwidth]{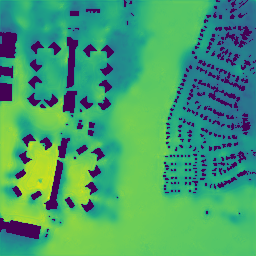} &
        \includegraphics[width=0.10\textwidth]{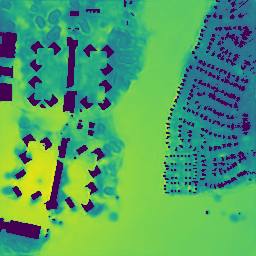} &
        \includegraphics[width=0.10\textwidth]{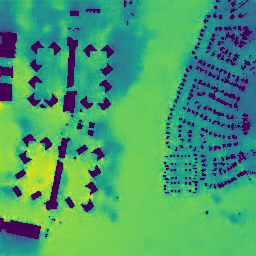} &
        \includegraphics[width=0.10\textwidth]{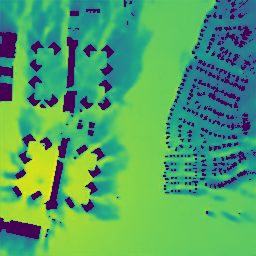} &
        \includegraphics[width=0.10\textwidth]{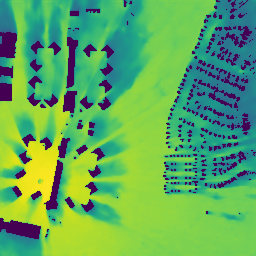} \\[-2pt]
        \scriptsize GT & \scriptsize KAN/31.85 & \scriptsize Kriging/25.43 & \scriptsize AE/42.71 & \scriptsize RadioUNet/15.03&
        \scriptsize C-GAN/18.82& \scriptsize RME-GAN/19.58 & \scriptsize Geo2Sig/14.30 & \scriptsize RadioKMoE/12.18 \\
    \end{tabular}

\caption{Comparison for predictions of multiband radiomaps ($2.4$\,GHz, $3.65$\,GHz, and $4.90$\,GHz) on the BRAT-LabW dataset at a 1\% sampling ratio. Each row corresponds to a different frequency, and values after the slash indicate MSE ($\times10^{-4}$).}

    \label{fig:bart_radiomap_results}
\end{figure*}

\begin{table*}[t]
\centering
\caption{$\overline{\text{NMSE}}/\overline{\text{MSE}}$ comparison on two datasets (NMSE $\times 10^{-3}$, MSE $\times 10^{-4}$).}
\label{table:result}
\begin{tabular}{lcccc|cccc}
\hline
\multirow{2}{*}{\textbf{Method}} & \multicolumn{4}{c|}{\textbf{BRAT-LabW Dataset (multiband)}} & \multicolumn{4}{c}{\textbf{RadioMapSeer Dataset (single-band)}} \\
\cline{2-5} \cline{6-9}
& \textbf{0.1\%} & \textbf{1\%} & \textbf{10\%} & \textbf{20\%} 
& \textbf{0.1\%} & \textbf{1\%} & \textbf{10\%} & \textbf{20\%} \\
\hline
KAN \cite{liu2025kan}            & 25.157/37.21 & 23.742/35.40 & 23.685/35.32 & 23.810/35.51  & 44.641/57.33 & 27.824/36.39 & 27.247/35.67 & 27.044/35.467 \\
Kriging\cite{kriging}         & 30.398/45.06  & 18.396/27.42 & 11.050/16.42 & 8.619/12.80  & 79.027/102.86 & 63.006/81.05 & 52.722/67.22 & 44.530/56.66 \\
AE \cite{AE}             & 26.726/40.91 & 28.928/44.26 & 14.738/22.59 & 13.955/21.39  & 21.729/27.55 & 7.541/9.61 & 4.599/5.89 & 5.668/7.26 \\
RadioUNet \cite{levie2021radiounet}      & 15.256/22.55 & 13.082/11.606 & 6.088/8.89 & 4.496/6.57  & 9.775/12.46 & 4.473/5.67 & 2.766/3.55 & 2.447/3.17 \\
cGAN \cite{mirza2014conditional}           & 20.254/29.63 & 14.562/21.41 & 7.137/10.44 & 5.141/7.47  & 10.562/13.55 & 4.092/5.22 & 2.802/3.62 & 3.010/3.98 \\
RME-GAN \cite{RME-GAN}        & 25.753/38.22  &  15.253/22.57 & 8.975 /13.24 & 7.355/10.87  & 32.259/40.76 & 5.413/6.96 & 3.214/4.15 & 3.367/4.34  \\
Geo2SigMap \cite{li2024geo2sigmap}      & 14.174/20.92 & 10.830 /15.89  & 5.274/7.67  & 4.172/6.08 & 7.233/9.11 & 2.598/3.34 & 1.003/1.28 & 1.062/1.39 \\
RadioKMoE& \textbf{12.352/16.68} & \textbf{9.845/14.00} & \textbf{4.920/7.14} & \textbf{3.324/4.82}  & \textbf{4.752/5.70} & \textbf{1.891/2.38} & \textbf{0.839/1.08} & \textbf{0.738/0.96} \\
\hline
\end{tabular}
\end{table*}

\section{Experiments}

\subsection{Experimental Setup}

\textbf{Dataset.}
We evaluate our framework on two datasets: BRAT-LabW and RadioMapSeer. Both datasets are represented as $256\times256$ images.

BRAT-LabW dataset contains 340 multiband radiomaps in a woodland-style (Woodland, CA, USA) urban scene at five carrier frequencies 
$\{2.40,\,3.65,\,4.90,\,5.80,\,28.00\}\,\mathrm{GHz}$ which is simulated by Sionna ray tracing \cite{sionna}, 
split into 300/20/20 for training/validation/testing. 
It targets frequency diversity under controlled propagation and evaluates performance in multiband RME.

RadioMapSeer \cite{DatasetPaper} provides 700 single-frequency radiomaps from metropolitan areas (Ankara, Berlin, Glasgow, Ljubljana, London, Tel Aviv). 
It contains a binary building map (building - 1; open space - 0). The transmitters are unit impulses, enabling precise localization and evaluation across heterogeneous urban morphologies.

\textbf{Metric.} We evaluate reconstruction quality using mean squared error (MSE) and normalized mean squared error (NMSE). 
Unless otherwise noted, metrics are computed on the linear (non-dB) scale over the full map and \textbf{averaged across frequencies}, i.e., $\overline{\text{MSE}}=\frac{1}{F}\sum\text{MSE}(f)$ and $\overline{\text{NMSE}}=\frac{1}{F}\sum\text{NMSE}(f)$ for $F$ frequencies, and
\begin{equation}
\mathrm{MSE}(f) = \frac{1}{HW}\,\big\| \widehat{R}_f - R_f \big\|_2^2; \quad \mathrm{NMSE}(f) = 
\frac{\big\| \widehat{R}_f - R_f \big\|_2^2}{\big\| R_f \big\|_2^2}.\nonumber
\end{equation}

\subsection{Overall Performance}
We compare our results with existing interpolation and learning approaches as shown in Table~\ref{table:result}. For each dataset, we report the $\overline{\text{NMSE}}/\overline{\text{MSE}}$ under four sampling ratios (0.1\%, 1\%, 10\%, 20\%). From the results,
classical interpolation (Kriging) \cite{kriging} shows reasonable trends as sampling increases, but remains sensitive to heterogeneous urban layouts. 
CNN baselines (AE\cite{AE}, RadioUNet\cite{levie2021radiounet}) improve over Kriging at moderate/high sampling yet degrade or become unstable at very sparse regimes (e.g., 0.1\%). 
Generative methods (cGAN\cite{mirza2014conditional}, RME-GAN\cite{RME-GAN}) narrow the gap in some settings but incur higher variance or training overhead.

Our method consistently achieves the lowest errors across all sampling ratios in both datasets. 
Improvements are most pronounced under sparse sampling (\textbf{0.1\%}--\textbf{1\%}), where the KAN prior captures the global propagation structure and the MoE refinement restores fine-grained detail near building edges and depth variations. 
The framework shows superior performance in accuracy and scalability at higher sampling ratios (\textbf{10\%} and \textbf{20\%}), across frequencies and diverse urban layouts. 
The results highlight the effectiveness of combining a data-driven KAN prior with region-wise MoE specialization for robust RME in complex environments.

\begin{figure}[t]
    \centering
    \setlength{\tabcolsep}{2pt} 
    \renewcommand{\arraystretch}{0.9} 
    \begin{tabular}{ccccc}
        \includegraphics[width=0.19\columnwidth]{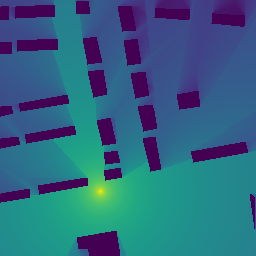} &
        \includegraphics[width=0.19\columnwidth]{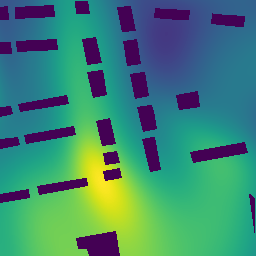} &
        \includegraphics[width=0.19\columnwidth]{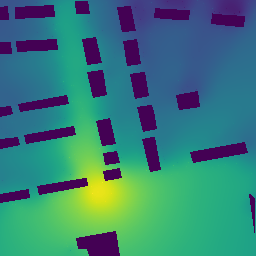} &
        \includegraphics[width=0.19\columnwidth]{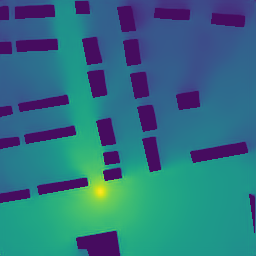} &
        \includegraphics[width=0.19\columnwidth]{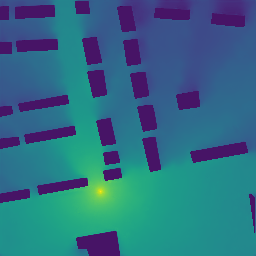} \\[-3pt]
        \scriptsize Ground Truth & \scriptsize KAN/36.39 & \scriptsize Kriging/81.05  & \scriptsize AE/9.61 & \scriptsize RadioUNet/5.67 \\[2pt]
        \includegraphics[width=0.19\columnwidth]{result/radiomapseer/gt.png} &
        \includegraphics[width=0.19\columnwidth]{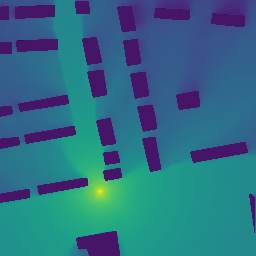} &
        \includegraphics[width=0.19\columnwidth]{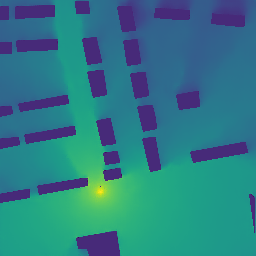} &
        \includegraphics[width=0.19\columnwidth]{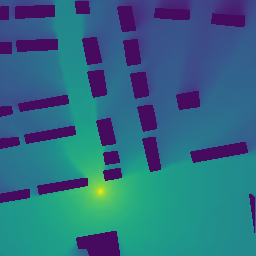} &
        \includegraphics[width=0.19\columnwidth]{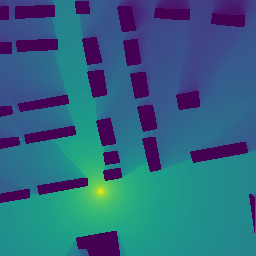} \\[-3pt]
        \scriptsize Ground Truth &  \scriptsize C-GAN/5.22& \scriptsize RME-GAN/6.96 & \scriptsize Geo2Sig/3.34 & \scriptsize RadioKMoE/2.38 \\
    \end{tabular}
    \caption{Comparison in the RadioMapSeer dataset at a 1\% sampling ratio. 
Values after the slash indicate MSE ($\times10^{-4}$).}

    \label{fig:radiomapseer_comparison}
\end{figure}

Fig.~\ref{fig:bart_radiomap_results} and Fig.~\ref{fig:radiomapseer_comparison} present visualization results on the two datasets.
In Fig.~\ref{fig:bart_radiomap_results}, each row corresponds to the BRAT-LabW dataset at three carrier frequencies (2.40~GHz, 3.65~GHz, and 4.90~GHz), respectively.
Our framework maintains consistently superior reconstruction quality across frequencies, especially displaying clear patterns of shadowing in the dense-building areas, demonstrating its ability to capture shared spatial patterns while adapting to frequency-dependent variations.
The KAN module produces a smooth coarse coverage prior that models large-scale propagation trends. The subsequent MoE refinement enhances fine details, preserves discontinuities near building boundaries, and reduces artifacts.
Fig.~\ref{fig:radiomapseer_comparison} shows the results on the single-frequency RadioMapSeer dataset, where similar improvements are observed in both structural accuracy and spatial consistency compared to other baselines.
 
\subsection{Ablation Study}



To evaluate the contribution of each module, we perform ablations at a 1\% sampling ratio on both BRAT-LabW and RadioMapSeer datasets.
We start from a TransUNet backbone and progressively add (i) the KAN-based coarse coverage prior (+KAN), (ii) MoE-based FFN replacement (+MoE), and (iii) a depth-map channel (+depth).
Table~\ref{tab:ablation} reports results of $\overline{\text{NMSE}}/\overline{\text{MSE}}$.
It shows that simply adding the KAN prior to the backbone does not effectively guide reconstruction.
When combined with the MoE module, however, the model achieves substantial performance gains over only +MoE and +KAN, as the expert specialization enhances the use of KAN’s coarse prior for fine-grained refinement.
Finally, incorporating the depth map yields the best overall results, where physics-inspired knowledge further improves the structural accuracy and robustness in complex urban environments.
\begin{table}[t]
\centering
\caption{Ablation study on BRAT-LabW and RadioMapSeer ($\overline{\text{NMSE}}/\overline{\text{MSE}}$) at 1\% sampling (NMSE $\times 10^{-3}$, MSE $\times 10^{-4}$).}
\label{tab:ablation}
\begin{tabular}{lcc}
\hline
\textbf{Backbone Variant} & \textbf{BRAT-LabW} & \textbf{RadioMapSeer} \\ 
\hline
Backbone                              & 11.349/16.66 & 2.734/3.54  \\
Backbone + KAN                        & 11.901/17.49 & 2.398/3.10  \\
Backbone + MoE                        & 10.347/15.15 & 2.275/2.93  \\
Backbone + MoE + KAN                  & \textbf{9.655}/14.12 & 2.058/2.66 \\
Backbone + MoE + KAN + Depth map      & {9.845}/\textbf{14.00} & \textbf{1.891}/\textbf{2.38} \\
\hline
\end{tabular}
\vspace{-5mm}
\end{table}

\section{Conclusion}
In this work, we propose a novel knowledge-guided framework for radiomap estimation that integrates KAN for global interpolation and MoE for adaptive fine-grained refinement. Specifically, the KAN module learns smooth coarse coverage priors directly from sparse measurements without heavy ray tracing or hand-crafted assumptions, providing a global trend estimation for the MoE refinement stage. 
The MoE leverages environmental and physics-inspired cues (building, transmitter, and depth maps) to adapt expert specialization across heterogeneous urban regions, enhancing local details while maintaining overall spatial consistency. 
Experiments on the BRAT-LabW and RadioMapSeer datasets demonstrate superior performance in both multiband and single-band radiomap estimation, validating the effectiveness of combining data-driven global priors with region-wise specialization in RME for next-generation spectrum management.


\bibliographystyle{IEEEtran} 
\bibliography{reference}

\end{document}